\documentclass[letterpaper, 10 pt, conference]{ieeeconf}  

\IEEEoverridecommandlockouts                              

\usepackage{amsmath,amssymb,amsfonts}
\usepackage{balance}
\usepackage{cite}
\usepackage{algorithmic}
\usepackage{graphicx}
\usepackage{textcomp}
\usepackage{xcolor}
\usepackage{csquotes}
\usepackage{siunitx}
\usepackage{subcaption}
\usepackage{gensymb}
\usepackage{url}
\usepackage{multirow}
\usepackage{colortbl}
\usepackage{wasysym}
\usepackage{booktabs}
\let\labelindent\relax
\usepackage{enumitem}
\usepackage{makecell}

\definecolor{mygray}{HTML}{434A52}
\definecolor{mygreen}{HTML}{7DB5A8}
\definecolor{myblue}{HTML}{6595BF}
\definecolor{myyellow}{HTML}{D6AE72}
\definecolor{myred}{HTML}{B75659}
\definecolor{failuremodetablecolor}{HTML}{B75659}

\makeatletter
\let\NAT@parse\undefined
\makeatother
\usepackage{hyperref}
\hypersetup{
    colorlinks,
    linkcolor=[HTML]{B75659},
    citecolor=[HTML]{B75659},
    urlcolor=[HTML]{6595BF}
}

\title{\LARGE \bf
 Instrumentation for Imitation Learning: Enhancing Training Datasets for Clothes Hanger Insertion
}

\author{Remko Proesmans$^{1}$, Thomas Lips$^{1}$ and Francis wyffels$^{1}$ 
\thanks{*This work was supported by the Research Foundation Flanders (FWO) under grant agreement no. 1S15925N and 1S56024N, and by the euROBIN Project (EU grant number 101070596). }
\thanks{$^{1}$Remko Proesmans, Thomas Lips and Francis wyffels are with the AI and Robotics Lab (IDLab-AIRO), Ghent University---imec, Ghent, Belgium
{\tt\small remko.proesmans@ugent.be}}%
}

\begin{document}

\maketitle
\thispagestyle{empty}
\pagestyle{empty}

\begin{abstract}

Large behaviour models have transformed the field of robotic manipulation, but prohibitive data requirements have thus far prevented a revolution similar to vision language models. 
We believe that instrumentation, i.e. sensor integration in objects, can provide invaluable state information and enable efficient learning for robotic manipulation.
In this paper, we present instrumented imitation learning of clothes hanger insertion.
Using 180 teleoperated demonstrations, we train diffusion policies with and without access to instrumentation data. 
Results show that policies leveraging instrumentation outperform vision-only counterparts by 14–25\,\%pt and exhibit greater task awareness. 
Crucially, a black-box imitation learning policy learns to prioritise instrumentation signals without explicit guidance. 
In addition, enhancing the teleoperation dataset with rollouts from an instrumented \enquote{expert} policy, enables a vision-only \enquote{student} policy to achieve performance comparable to the instrumented expert, thereby surpassing the original vision-only policy. 
These findings establish instrumentation as a promising strategy to enhance imitation learning for robotic manipulation.
Datasets are available on Zenodo [\href{https://doi.org/10.5281/zenodo.17122216}{10.5281/zenodo.17122216}].
\end{abstract}

\section{INTRODUCTION}

Great expectations currently surround the development of large behaviour models (LBMs) for robotics because of the recent success of vision-language models (VLMs) in image and text generation, or software-based environments in general~\cite{billard2025, goldberg2025, barreiros2025TRI-LBM}.
However, it remains to be seen whether these techniques can transfer to real-time sensing, planning, and control for physical machines operating in unpredictable, human environments~\cite{billard2025}.
At the very least, we are far from reaching the internet-scale data requirements to make them feasible~\cite{goldberg2025}.
A popular way to achieve advanced manipulation skills for robots is to make them imitate human behaviour through demonstrations, called imitation learning (IL)~\cite{ARGALL2009469}. 
This has proven to be effective in teaching complex manipulation skills to robots \cite{act, chi2024diffusionpolicy, pi0, openvla}.
However, learning even a single task in a controlled environment can take 100 or more demonstrations~\cite{act, chi2024diffusionpolicy}.
For a robot to handle multiple complex tasks in unstructured environments, the data requirements grow tremendously. 

Large datasets for robot pre-training exist~\cite{openxembodiment, droidinthewilddata, bridgedatav2datasetrobot}, but current LBMs still require task-specific fine-tuning for complex tasks or diverse environments~\cite{pi0}.
Obtaining larger datasets can be done by adding successful policy rollouts to the training set and retraining, but this merely biases the data distribution
toward states where the policy is already proficient~\cite{mirchandani2024thinkscaleautonomousrobot}.
Alternatively, in interactive imitation learning~\cite{celemin2022} expert interventions can get a struggling policy rollout back on track~\cite{ingelhag2025realtimeoperatortakeovervisuomotor, kelly2019hgdaggerinteractiveimitationlearning, dai2024racerrichlanguageguidedfailure}.
This allows for deliberate data collection to efficiently overcome data mismatches between the train data and deployment settings~\cite{kelly2019hgdaggerinteractiveimitationlearning}.

In addition to larger datasets and algorithmic advances, we can attempt to improve the quality of the demonstrations to increase IL capabilities.
Both~\cite{maram2022} and \cite{bilal2024} determined evaluation metrics for the quality of demonstration trajectories, and showed that better demonstrations can result in a better policy for the same amount of data.
Xu et al.~\cite{xu2024rldg} have shown that reinforcement learning (RL) agents generate high-quality trajectories through reward maximisation, making them better suited for fine-tuning generalist policies compared to human demonstrations.

\begin{figure}
    \centering
    \includegraphics[width=\linewidth]{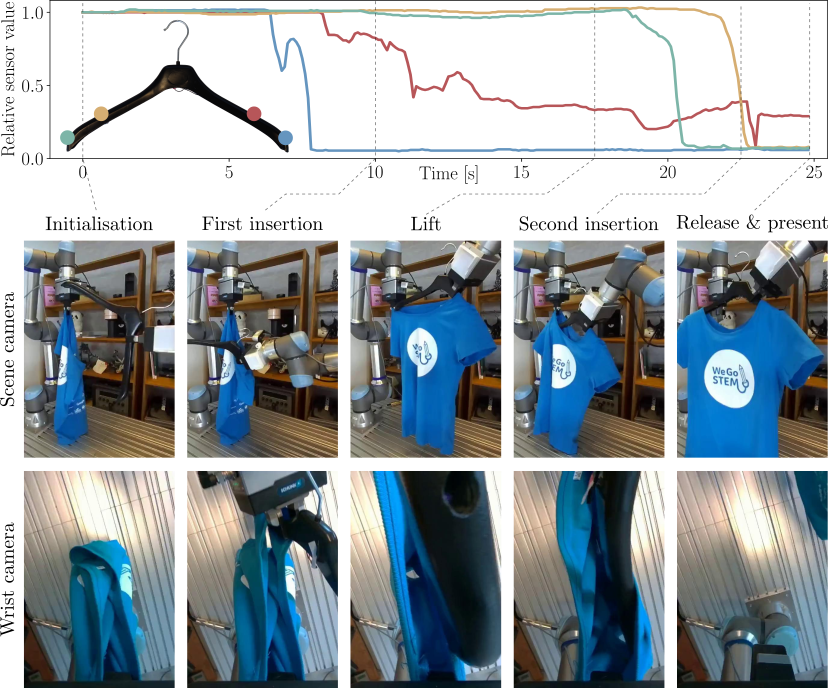}
    \caption{Instrumented clothes hanger insertion.}
    \label{fig:task_overview}
\end{figure}

In~\cite{Junge2023}, a different approach from standard IL was used for raspberry harvesting: instead of teleoperating the robot and recording trajectories as demonstrations, human demonstrators manipulated an instrumented, i.e. sensorised, strawberry phantom.
Then, the parameters of a custom controller were automatically tuned to match human behaviours as experienced by the strawberry.
Instrumentation was also used in~\cite{openai2019} for learning to solve a Rubik's cube with an anthropomorphic robot hand. 
Specifically, the sensor data provided accurate face angle observations of the Rubik's cube, adding to the state information needed to shape the reward function.
However, when moving to a vision-only policy, the sensorised cube was discarded and replaced by a synthetically trained vision model: the sensorised cube was rather a stepping stone in development, than a critical influence in obtaining a functional policy.
Similarly, \cite{verleysen2020} and \cite{smarttextile} use sensors for state estimation of cloth to enable robotic folding, but do not yet transfer to garments without sensors.

The fundamental idea of instrumentation is to use additional state information during the learning phase and then deploy without this privileged information, using only sensory input that is available in the field.
This concept is also found in works that learn robot behaviours in simulation~\cite{lee2020,chen2023}. 
In~\cite{lee2020}, sim-to-real quadrupedal locomotion is achieved by first training a teacher policy that has access to privileged state information of the robot, allowing it to quickly achieve high performance. 
The teacher is then used to guide the learning of a student controller that only uses sensors available on the real robot.
Similarly in~\cite{chen2023}, the student-teacher paradigm is used to learn in-hand reorientation.
The teacher used the object orientation to quickly learn appropriate actions, after which the student was trained to match the teacher's outputs while relying on occluded visual observations of the objects.
Instrumentation like in \cite{Junge2023, openai2019, verleysen2020, smarttextile} can be a way to obtain similar privileged information directly in the real world, potentially allowing for faster learning and/or more performant policies and avoiding the need for sim-to-real transfer, which already is a significant challenge on its own. 

This work presents foundational results uncovering the potential of instrumentation for IL through a case study focusing on the task of clothes hanger insertion.
We show that (1)~a black-box IL policy can recognise the importance of instrumentation signals without explicit guidance, (2)~including instrumentation in the training data for an IL policy can increase both the policy's task awareness as well as overall performance, and (3)~that rollouts from such an \enquote{expert} policy can be used as extra training data to enhance the performance of a \enquote{student} policy that cannot rely on instrumentation.

\section{Materials \& Methods}

\subsection{Task Description}
Cloth manipulation is a challenging research topic in robotics requiring advanced methods for modelling, perception, and control~\cite{longhini2025unfolding}. 
The field comprises a variety of subtasks, like collection, unfolding, folding, hanging, etc.
Whereas many systems have been proposed for, e.g., unfolding and folding, the task of clothes hanger insertion has received limited attention~\cite{robohanger2025, koishihara2017}.
We choose clothes hanger insertion as a testbed for instrumentation, because the (mostly) rigid clothes hanger lends itself to sensor integration, while the task as a whole is challenging and rarely addressed in the state-of-the-art of cloth manipulation.

Fig.~\ref{fig:task_overview} shows the progression of the insertion task.
In order to alleviate data requirements and not lose focus from the core research questions of this work regarding the potential of instrumentation, some constraints apply to task initialisation: 
\begin{itemize}
    \item Both the T\nobreakdash-shirt and the clothes hanger are already held by the robot arms.
    \item We only use one T\nobreakdash-shirt and one clothes hanger, always held in the same location.
\end{itemize}
Once started, the robot inserts the first leg of the hanger in the open collar of the T\nobreakdash-shirt, and lifts the clothes hanger until it is in a good position to perform the insertion of the second leg.
After the second insertion, the T\nobreakdash-shirt is released so that it hangs only from the clothes hanger.
For evaluation rollouts, \enquote{task success} is defined as the autonomous execution of these stages, concluding with the T-shirt hanging only from the clothes hanger, on both shoulders.

\subsection{Hardware setup}

\subsubsection{Instrumented Clothes Hanger}

\begin{figure}[t]
\centering
\begin{subfigure}[t]{0.5\textwidth}
    \centering
    \includegraphics[width=0.7\linewidth]{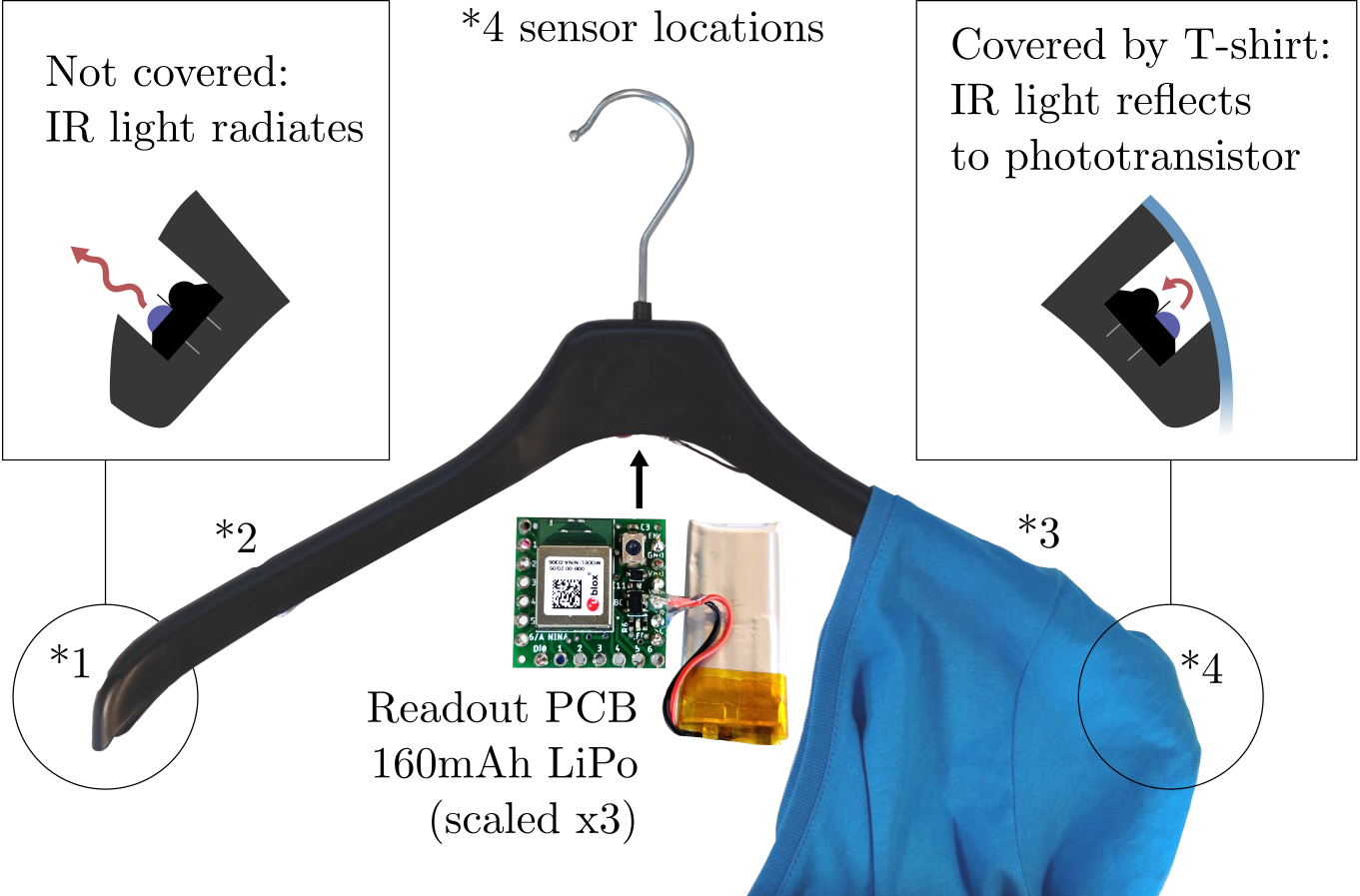}
    \caption{Sensor integration and working principle. The electronics fit inside the hollow clothes hanger.}
    \label{fig:clothes_hanger_annotated}
\end{subfigure}\hfill
\vspace{2mm}
\begin{subfigure}[t]{0.5\textwidth}
    \centering
    \includegraphics[width=0.9\linewidth]{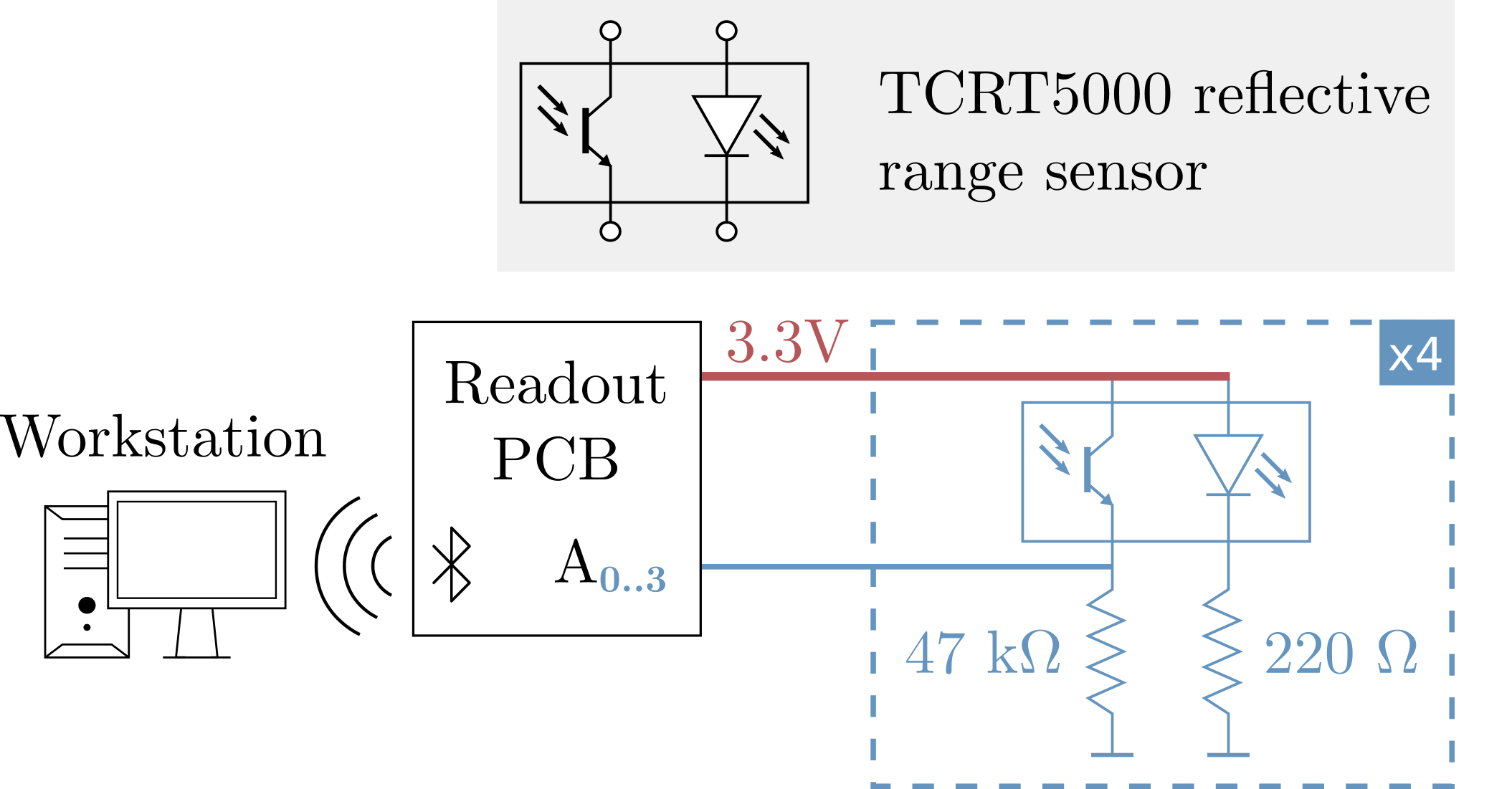}
    \caption{Electrical schematic.}
    \label{fig:schematic}
\end{subfigure}
\caption{Instrumented clothes hanger with integrated infrared (IR) reflective range sensors. The range sensors give a strongly different signal when the clothes hanger is covered, compared to when it is not.}
\label{fig:instrumented_clothes_hanger}
\end{figure}

A standard clothes hanger is instrumented by integration of four TCRT5000 reflective infrared (IR) range sensors, see Fig.~\ref{fig:instrumented_clothes_hanger}.
The TCRT5000 contains an IR light-emitting diode (LED), and an IR phototransistor (PT).
When the clothes hanger is not covered, the IR light radiates away, but if the T\nobreakdash-shirt covers it, the light reflects back to the PT, causing a large change in the photocurrent, as can be seen in the top curves of Fig.~\ref{fig:task_overview}.
For readout, we use a printed circuit board (PCB) from our open-source, modular, wireless Smart Textile~\cite{smarttextile}, featuring an nRF52840 microcontroller.
The PCB, as well as a 160\,mAh LiPo battery, easily fit inside the hollow clothes hanger. 
Data is communicated to a workstation over Bluetooth Low Energy (BLE).

\subsubsection{Robot Setup}
\begin{figure}[t]
\centering
\begin{subfigure}[t]{0.5\textwidth}
    \centering
    \includegraphics[width=0.8\linewidth]{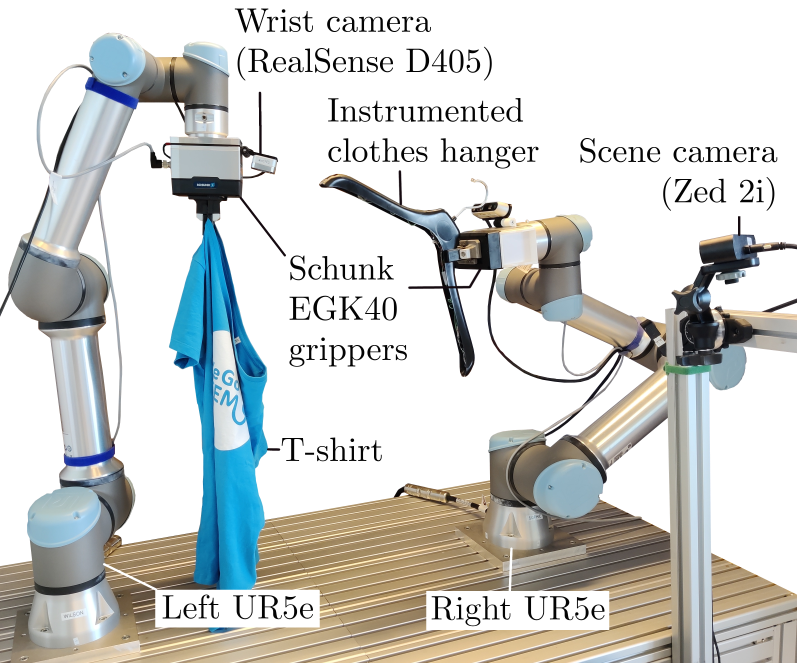} 
    \caption{Robot setup.}
    \label{fig:setup}
\end{subfigure}\hfill
\vspace{1mm}
\begin{subfigure}[t]{0.5\textwidth}
    \centering
    \includegraphics[height=3.3cm]{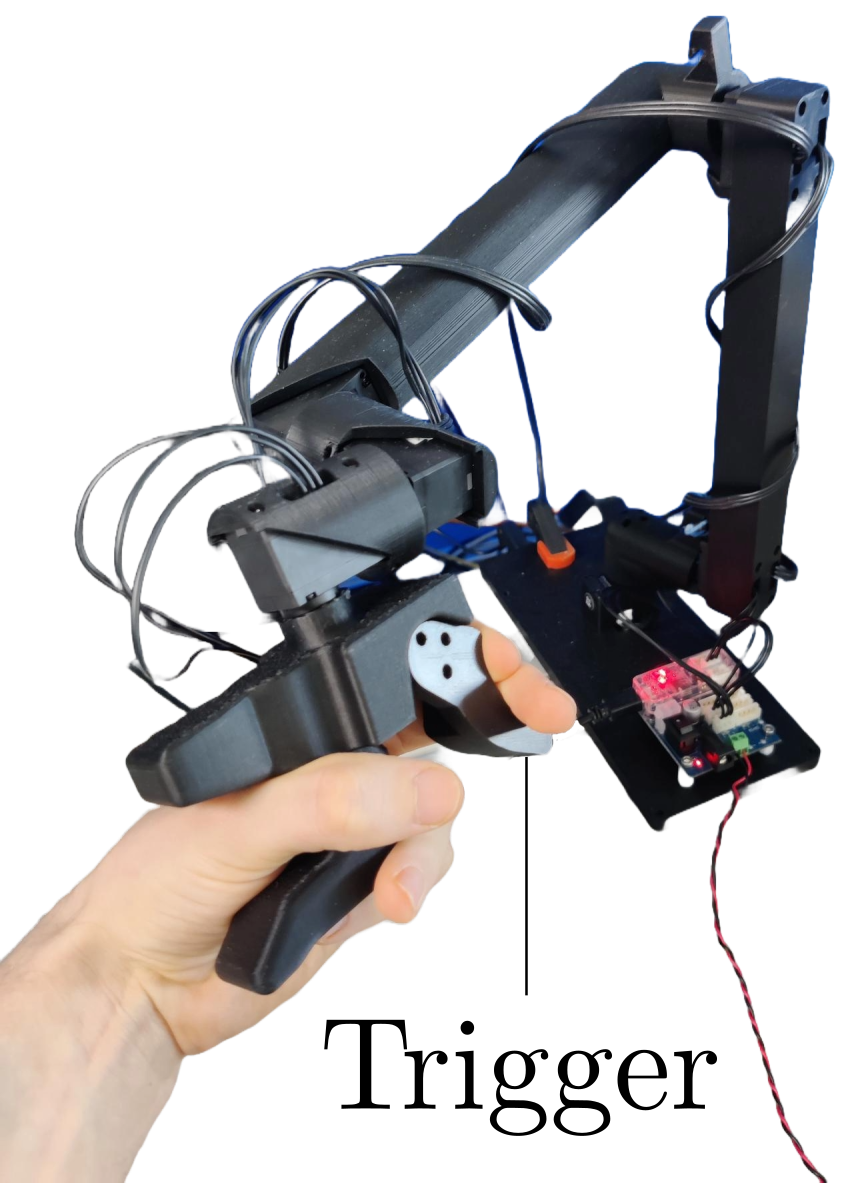}
    \caption{The Gello \cite{gello} teleoperation arm.}
    \label{fig:teleop_closeup}
\end{subfigure}
\caption{Experimental setup for learning clothes hanger insertion.}
\label{fig:experimental_setup}
\end{figure}

The robot system is shown in Fig.~\ref{fig:setup}.
It features two UR5e collaborative robot arms: the left holds the T\nobreakdash-shirt, the right moves the clothes hanger.
A Schunk EGK\nobreakdash-40\nobreakdash-MB\nobreakdash-M\nobreakdash-B gripper is mounted to each arm. 
Standard Robotiq 2F\nobreakdash-85 fingertips are fitted to the Schunk grippers by means of 3D-printed mechanical adapters.
The left robot arm has a RealSense~D405 camera attached to its wrist to get a close-up view of the collar opening of the T\nobreakdash-shirt.
An externally mounted Zed~2i camera provides a view of the entire scene.

\subsubsection{Teleoperation Setup}
We use a Gello~\cite{gello} arm (Fig.~\ref{fig:teleop_closeup}) to control the joint positions of the right UR5e, the left UR5e is static.
The Gello arm trigger is configured to control the opening of the Schunk gripper on the left UR5e, the opening of the Schunk gripper on the right UR5e is fixed.

\subsection{Policy Architecture} \label{ss:learningarch}

For the policies trained in this work, we use a Diffusion Policy~\cite{diffusion} architecture with a ResNet18 vision backbone and a CNN-based noise prediction network. We set the hyperparameters to the default values from~\cite{diffusion} unless specified otherwise.
The model predicts the joint positions of the right UR5e and the gripper opening of the left UR5e, given an observation that consists of 720x500x3 images from both wrist and scene camera, and a state vector.
The state vector is composed of the six joint poses of the right UR5e, the gripper opening of the left UR5e, and optionally the four sensor readings of the clothes hanger: in the following, policies with and without access to instrumentation data are compared.
The latter may be denoted as \enquote{vision-only} policies, even though the robot joint poses and gripper opening are still included in the state.
The model predicts a chunk of 16 actions, of which 8 are executed before predicting another chunk.
At a control frequency of 10\,Hz, these 8 actions correspond to 0.8\,s of wall-clock time.
We train all models for 100\,000 steps with a batch size of 32, which  takes about 11\,h on an NVIDIA RTX4090 GPU, inference takes 270\,ms. 

In this work, a total of seven policies are trained.
Policies for which the clothes hanger sensor data is included in the state vector are denoted as $\Pi^t_{\textrm{instr}}$, vision-only policies as $\Pi^t_{\textrm{vis}}$, with $t$ being the number of demonstrations that the policy was trained on.

\subsection{Data Collection}

Three different types of datasets are collected in this work, each following their own protocol: (1) human demonstrations are collected through teleoperation (subsection~\ref{sss:teleop}); (2) policy rollouts are collected to estimate success rates (subsection~\ref{sss:eval}); (3) policy rollouts from a policy relying on instrumentation are added to the teleoperation data, forming an enhanced, larger dataset to train better vision-only policies (subsection~\ref{sss:aug}).
A single episode (demo or rollout) can take four forms:
\begin{itemize}
    \item[] \textbf{Type I:} Initialisation exactly as depicted in Fig.~\ref{fig:task_overview} and \ref{fig:setup}, with the right UR5e starting from a fixed home pose. The task is executed to completion.
    \item[] \textbf{Type II:} The right arm starts from a \enquote{missed first insertion}, i.e., the first leg of the clothes hanger is near to, but below the level of, the collar of the T\nobreakdash-shirt. The task is executed to completion.
    \item[] \textbf{Type III:} The right arm starts from a \enquote{missed first insertion}, the episode ends right after the first insertion.
    \item[] \textbf{Type IV:} The right arm starts from its home pose, the episode ends right after the first insertion.
\end{itemize}
While Type~I represents the full task, Types~II, III and IV are focused on achieving the first insertion, as this stage of the task was found to be much more difficult than the second insertion, see the failure mode analysis in subsection~\ref{ss:failure_mode}.
The initial state of the T\nobreakdash-shirt is randomised by manually lifting the shoulder that is not grasped by the robot and letting it go.
In section~\ref{s:results}, Fig.~\ref{fig:distributions}, we will show that this procedure provided sufficient randomisation.
Datasets are available on Zenodo [\href{https://doi.org/10.5281/zenodo.17122216}{10.5281/zenodo.17122216}].


\subsubsection{Teleop training data} \label{sss:teleop}
In total, 180 human demonstrations of the task shown in Fig.~\ref{fig:task_overview} are collected. Three subsets of this data are defined:
\begin{itemize}
    \item[] \textbf{\#train=180:} This is the full teleop set, comprising 50 Type~I demonstrations, 120 Type~II demonstrations, 10 Type~III demonstrations. 
    \item[] \textbf{\#train=100:} 20 Type~I demonstrations, 80 Type~II demonstrations. 
    \item[] \textbf{\#train=50:} 50 Type~II demonstrations. 
\end{itemize}
The \textbf{\#train=180} set comprises 47\,040 observations, which accounts for 1.31\,h of data collection. However, this excludes resetting the environment between demonstrations, compressing and saving each demonstration, and occasionally reinitialising the entire data collection pipeline after e.g. a collision. 
Overall, we estimate a data collection time of 3.5\,h.

\subsubsection{Policy rollouts for evaluation metrics} \label{sss:eval}
All policy rollouts used to calculate success rates are of Type I.
They are given at least one minute to complete an insertion, both for the first and second.
If after one minute the policy is not reaching new states (anecdotally, this mostly occurs when the clothes hanger is not touching the T\nobreakdash-shirt), the rollout is ended and declared a failure.
If the policy takes longer than two minutes for an insertion, the rollout is ended no matter how close it is to completion.

\subsubsection{Policy rollouts for enhanced dataset} \label{sss:aug}
In the Results section~(\ref{s:results}), it will be shown that including instrumentation data increases both a policy’s task awareness and overall performance.
However, policies relying on instrumentation cannot be deployed directly in the field since we do not envision every household object to have embedded sensors: we need ways to omit the instrumentation data without losing performance.
In this work, we use rollouts from a policy relying on instrumentation as extra training data to obtain a more performant vision-only policy.

Specifically, we add the successful evaluation rollouts (Type~I) of $\Pi^{180}_{\textrm{instr}}$ to the \textbf{\#train=180} set.
In addition, we review the failures of $\Pi^{180}_{\textrm{vis}}$ and recreate their initialisations as closely as possible by overlaying their initial camera observations with a live camera feed.
Then, we let $\Pi^{180}_{\textrm{instr}}$ attempt Type~IV rollouts on each of the difficult initialisations, until five successful demonstrations are obtained, with a maximum of ten attempts per initialisation.
As such, we obtain an enhanced dataset, \textbf{\#train=237}, for training a better vision-only policy $\Pi^{237}_{\textrm{vis+}}$ (the extra \enquote{+} subscript emphasises the enhanced training data compared to other $\Pi_{\textrm{vis}}$ policies).
Note that we do not filter the $\Pi^{180}_{\textrm{instr}}$ rollouts for optimal behaviour: mistakes and corrective behaviours are included as long as the rollout is successful.
See subsection~\ref{ss:policy_success} for more details on the additional 57 trials.

\section{Results} \label{s:results}

\begin{figure}
    \centering
    \includegraphics[width=\linewidth]{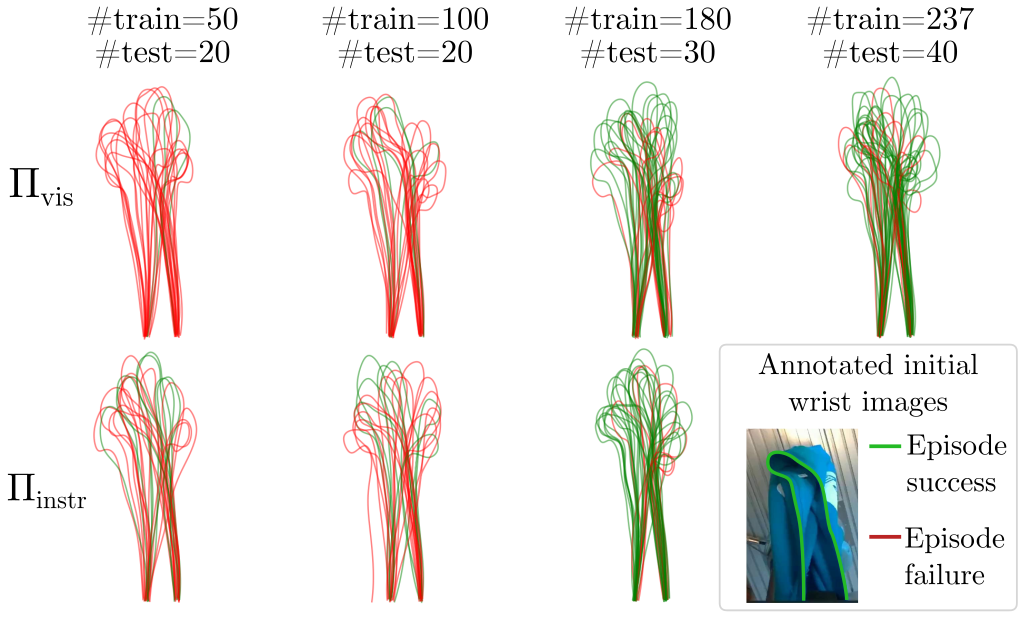}
    \caption{Visualisation of each test set. The curves are traces of the collar of the T-shirt as seen from the wrist camera at the start of an episode.}
    \label{fig:distributions}
\end{figure}

Seven policies are trained: $\Pi^{50}_{\textrm{instr}}$ and $\Pi^{50}_{\textrm{vis}}$ on the \textbf{\#train=50} set, $\Pi^{100}_{\textrm{instr}}$ and $\Pi^{100}_{\textrm{vis}}$ on the \textbf{\#train=100} set, $\Pi^{180}_{\textrm{instr}}$ and $\Pi^{180}_{\textrm{vis}}$ on the \textbf{\#train=180} set, and $\Pi^{237}_{\textrm{vis+}}$ on the enhanced \textbf{\#train=237} set.
$\Pi^{50}$ and $\Pi^{100}$ policies are evaluated with 20 rollouts each, $\Pi^{180}$ policies with 30, and the $\Pi^{237}_{\textrm{vis+}}$ policy with 40 rollouts.
The initial states of the evaluation sets are visualised in Fig.~\ref{fig:distributions}.
This figure overlays the collar of the T\nobreakdash-shirt as seen from the wrist camera at the start of each rollout for each of the evaluation sets.
It indicates that the distribution of initial states is sufficiently random, and that failures and successes are not biased towards any type of initial state.

\subsection{Policy success rates} \label{ss:policy_success}

\begin{figure}
    \centering
    \includegraphics[width=\linewidth]{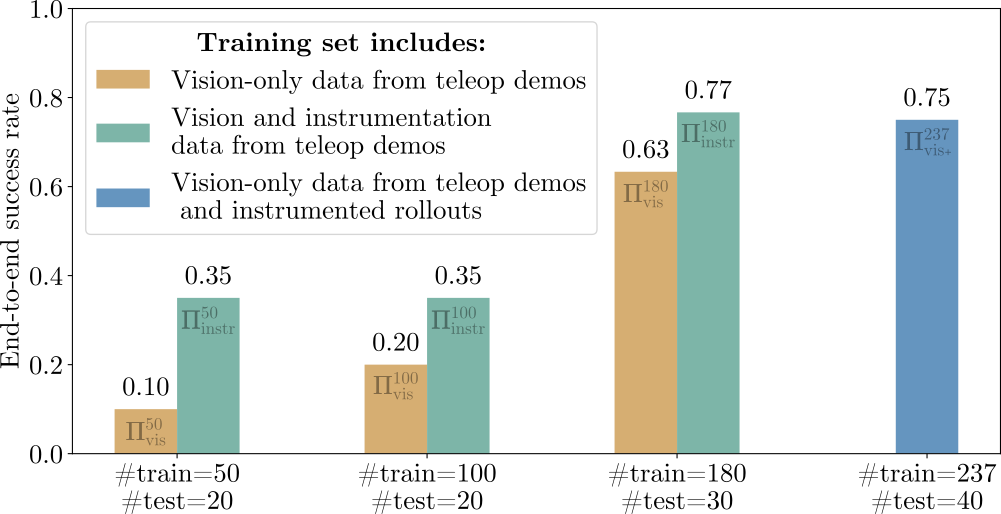}
    \caption{End-to-end success rates for diffusion policies trained on different dataset sizes, including different modalities. Instrumented policies outperform their vision-only counterparts. A vision-only policy can be improved by adding rollouts from an instrumented policy to the training data.}
    \label{fig:results}
\end{figure}

The experimental end-to-end success rates are shown in Fig.~\ref{fig:results}.
For a given training set size, a policy with access to instrumentation data consistently outperforms a vision-only policy by 14 to 25\,\%pt.
All 23 successful $\Pi^{180}_{\textrm{instr}}$ rollouts from the test set, as well as 34 additional $\Pi^{180}_{\textrm{instr}}$ rollouts targeting the 11 $\Pi^{180}_{\textrm{vis}}$ failures according to the formalism described in subsection~\ref{sss:aug}, are added to the \textbf{\#train=180} set to obtain the \textbf{\#train=237} set.
The vision-only policy $\Pi^{237}_{\textrm{vis+}}$ trained on \textbf{\#train=237} outperforms $\Pi^{180}_{\textrm{vis}}$ by 12\,\%pt.
To better frame these results, we perform a Bayesian analysis~\cite{kressgazit2024robotlearningempiricalscience}.
Task success is treated as a Bernoulli distribution with unknown parameter $p$, the probability of success, which we can estimate given the experimental results.
According to Bayes' theorem, the probability density function (PDF) of $p$ given $s$ measured successes out of $N$ rollouts, P$(p|s)$, can be expressed as:
\begin{equation} \label{e:pdf_successrate}
    \textrm{P}(p|s)=\textrm{P}(s|p)\frac{\textrm{P}(p)}{\textrm{P}(s)}.
\end{equation}
Assuming a uniform distribution on $p$, P$(p)$ is equal to 1.
P$(s|p)$, the PDF of the number of successes when running $N$ Bernoulli experiments given success rate $p$, follows a binomial distribution:
\begin{equation} \label{e:binomial}
    \textrm{P}(s|p)=p^s(1-p)^{N-s}\binom{N}{s}.
\end{equation}
Lastly, P$(s)$ is calculated as:
\begin{equation}
\begin{split}
    \textrm{P}(s) & =\int_0^1\textrm{P}(s, p)\textrm{d}p =\int_0^1\textrm{P}(s|p)\textrm{P}(p)\textrm{d}p \\
    & =\int_0^1\textrm{P}(s|p)\textrm{d}p\stackrel{(\ref{e:binomial})}{=}\binom{N}{s}\int_0^1p^s(1-p)^{N-s}\textrm{d}p \\
    & =\binom{N}{s}\textrm{B}(s+1, N-s+1)\\
    & =\binom{N}{s}\frac{\Gamma(s+1)\Gamma(N-s+1)}{\Gamma(N+2)}\\
    & = \frac{N!}{s!(N-s)!}\frac{s!(N-s)!}{(N+1)!}=\frac{1}{N+1}.
\end{split}
\end{equation}
with B and $\Gamma$ the well-known Beta and Gamma functions, respectively. We can now write equation~\ref{e:pdf_successrate} as:
\begin{equation}
    \textrm{P}(p|s)=p^s(1-p)^{N-s}\frac{(N+1)!}{s!(N-s)!}.
\end{equation}
It follows that P$(p|s)\sim\textrm{Beta}(s+1, N-s+1)$, or, with $\hat{p}$ the experimentally determined success rate as reported in Fig.~\ref{fig:results}: P$(p|\hat{p})\sim\textrm{Beta}(N\hat{p}+1, N(1-\hat{p})+1)$.
This way, we can estimate the distribution of $p$ for each $\Pi$.
Using Monte Carlo simulation, we obtain:

\begin{equation} \label{e:bayesian_success_rates}
\left\{
\begin{aligned}
    &\textrm{P}[p^{50}_{\textrm{instr}}\sim\textrm{Beta}(8, 14)>p^{50}_{\textrm{vis}}\sim\textrm{Beta}(3, 19)] = 96.7\,\% \\
    &\textrm{P}[p^{100}_{\textrm{instr}}\sim\textrm{Beta}(8, 14)>p^{100}_{\textrm{vis}}\sim\textrm{Beta}(5, 17)] = 84.7\,\% \\
    &\textrm{P}[p^{180}_{\textrm{instr}}\sim\textrm{Beta}(24,8)>p^{180}_{\textrm{vis}}\sim\textrm{Beta}(20,12)] = 86.5\,\% \\
    &\textrm{P}[p^{237}_{\textrm{vis}}\sim\textrm{Beta}(31,11)>p^{180}_{\textrm{vis}}\sim\textrm{Beta}(20,12)] = 85.3\,\%.
\end{aligned}
\right.
\end{equation}

\subsection{Failure Modes} \label{ss:failure_mode}

\begin{figure}[t]
\centering
\vspace{2.7mm}
\begin{subfigure}[t]{0.48\textwidth}
    \centering
    \includegraphics[width=0.8\linewidth]{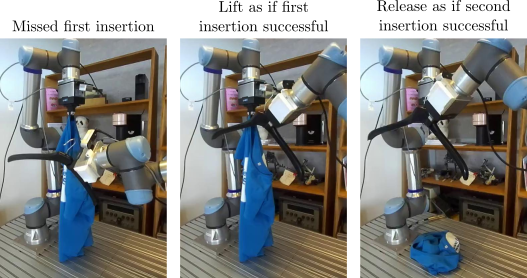}
    \caption{A vision-only policy often executes robot movements without concern for the state of the T\nobreakdash-shirt, resulting in a dropped T\nobreakdash-shirt.}
    \label{fig:not_aware_n100}
\end{subfigure}\hfill
\vspace{2mm}
\begin{subfigure}[t]{0.48\textwidth}
    \centering
    \includegraphics[width=0.8\linewidth]{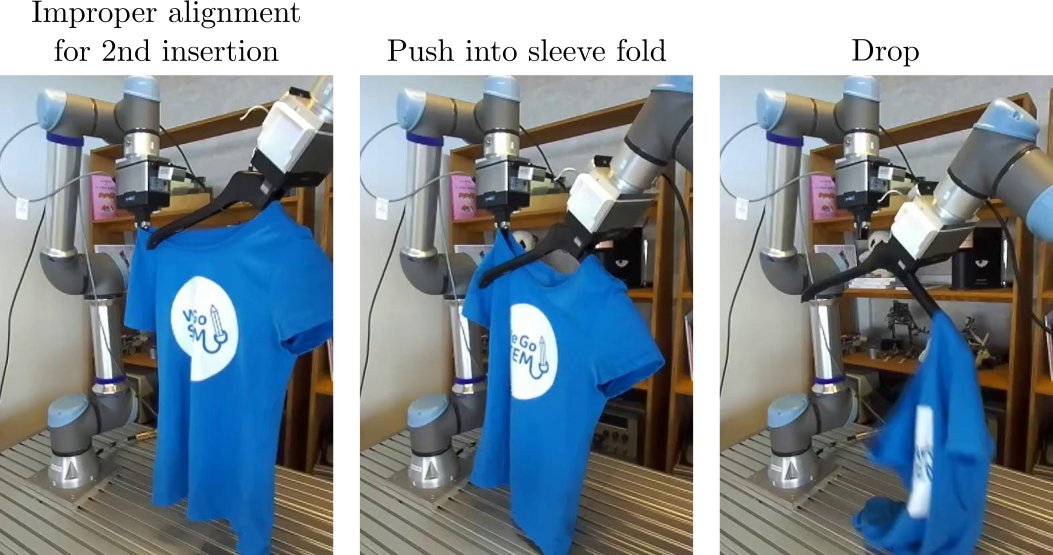}
    \caption{To an instrumented policy, a sleeve draped over the clothes hanger can look similar to a properly inserted hanger.}
    \label{fig:2nd_insert_drop}
\end{subfigure}\hfill
\vspace{2mm}
\begin{subfigure}[t]{0.48\textwidth}
    \centering
    \includegraphics[width=0.8\linewidth]{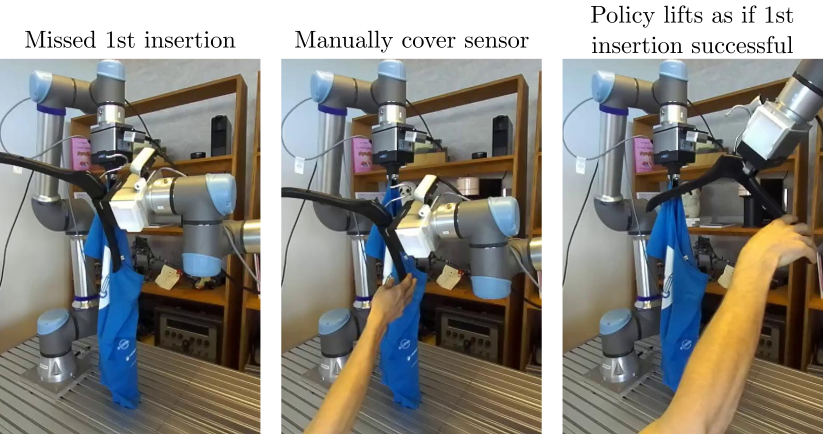}
    \caption{An instrumented policy relies more strongly on the sensor data than the visual feed. After a failed first insertion, the policy realises it cannot continue. When manually covering the sensor, the policy attempts to lift, disregarding visual cues.}
    \label{fig:instrumentation_cue}
\end{subfigure}
\caption{Policy errors and behaviours related to task awareness.}
\label{fig:awareness_n100}
\end{figure}

We distinguish the following failure modes for unsuccessful policy rollouts:
\begin{itemize}
    \item[] \textbf{Collision\,[1st insertion]:} A collision occurs between clothes hanger, gripper, and/or wrist camera before a successful first insertion.
    \item[] \textbf{Drop\,[1st insertion]:}  The gripper of the left UR5e opens and drops the T\nobreakdash-shirt before a successful first insertion has been performed (see Fig.~\ref{fig:not_aware_n100}).
    \item[] \textbf{Stuck\,[1st insertion]:} The policy attempts the first insertion, but misses the collar of the T\nobreakdash-shirt and gets stuck.
    \item[] \textbf{Pulled drop:}  The policy attempts the first insertion, but pulls the T\nobreakdash-shirt down and out from between the fingertips of the left UR5e.
    \item[] \textbf{Failed lift:} The policy succeeds in the first insertion, but does not move on to lift the T\nobreakdash-shirt.
    \item[] \textbf{Stuck\,[2nd insertion]:} The policy gets stuck right before the second insertion.
    \item[] \textbf{Collision\,[2nd insertion]:} A collision occurs between clothes hanger, gripper, and/or wrist camera during an attempt of the second insertion.
    \item[] \textbf{Drop\,[2nd insertion]:}  The gripper of the left UR5e opens and drops the T\nobreakdash-shirt after an unsuccessful attempt at the second insertion (see Fig.~\ref{fig:2nd_insert_drop}).
\end{itemize}

Table~\ref{tab:failure_modes} compiles the incidence of each failure mode for each of the seven policies.
The colour scale is normalised with respect to the number of failures in the test set for each policy.

\begin{table*}[]
\centering
\caption{Policy failure mode incidence (color scale normalised per row).}
\label{tab:failure_modes}
\setlength{\tabcolsep}{2pt}
\renewcommand{\arraystretch}{1.5} 
\begin{tabular}{c|c|c|cccccccc}
   Policy & \makecell{\#\,Rollouts \\ in test set} & \makecell{\#\,Failures \\ in test set}
   & \makecell{\textbf{Collision} \\ \textbf{[1st insertion]}} 
   & \makecell{\textbf{Drop} \\ \textbf{[1st insertion]}} 
   & \makecell{\textbf{Stuck} \\ \textbf{[1st insertion]}}
   & \makecell{\textbf{Pulled} \\ \textbf{drop}}
   & \makecell{\textbf{Failed} \\ \textbf{lift}}
   & \makecell{\textbf{Stuck} \\ \textbf{[2nd insertion]}} 
   & \makecell{\textbf{Collision} \\ \textbf{[2nd insertion]}} 
   & \makecell{\textbf{Drop} \\ \textbf{[2nd insertion]}}  \\
  \hline
  $\Pi^{50}_{\textrm{vis}}$ & 20 & 18 & \cellcolor{failuremodetablecolor!11}2 & \cellcolor{failuremodetablecolor!17}3 & \cellcolor{failuremodetablecolor!61}11 & 0 & 0 & 0 & 0 & \cellcolor{failuremodetablecolor!11}2\\
  $\Pi^{50}_{\textrm{instr}}$ & 20 & 13 & 0 & 0 & \cellcolor{failuremodetablecolor!69}9 & \cellcolor{failuremodetablecolor!8}1 & 0 & 0 & 0 & \cellcolor{failuremodetablecolor!23}3\\

  \hline

  $\Pi^{100}_{\textrm{vis}}$ & 20 & 16 & 0 & \cellcolor{failuremodetablecolor!44}7 & \cellcolor{failuremodetablecolor!38}6 & 0 & 0 & \cellcolor{failuremodetablecolor!6}1 & \cellcolor{failuremodetablecolor!13}2 & 0 \\
  $\Pi^{100}_{\textrm{instr}}$ & 20 & 13 & 0 & 0 & \cellcolor{failuremodetablecolor!46}6 & 0 & 0 & \cellcolor{failuremodetablecolor!46}6 & \cellcolor{failuremodetablecolor!6}1 & 0 \\
  
  \hline
  
  $\Pi^{180}_{\textrm{vis}}$ & 30 & 11 & 0 & 0 & \cellcolor{failuremodetablecolor!100}11 & 0 & 0 & 0 & 0 & 0 \\
  $\Pi^{180}_{\textrm{instr}}$ & 30 & 7 & 0 & 0 & \cellcolor{failuremodetablecolor!43}3 & \cellcolor{failuremodetablecolor!43}3 & 0 & \cellcolor{failuremodetablecolor!14}1 & 0 & 0 \\
  
  \hline
  
  $\Pi^{237}_{\textrm{vis+}}$ & 40 & 10 & 0 & 0 & \cellcolor{failuremodetablecolor!70}7 & \cellcolor{failuremodetablecolor!10}1 & \cellcolor{failuremodetablecolor!10}1 & 0 & \cellcolor{failuremodetablecolor!10}1 & 0 \\
\end{tabular}
\end{table*}

\section{Discussion}

The experimental success rates in Fig.~\ref{fig:results} show that the instrumented policies consistently outperform their vision-only counterparts.
The gap is largest (25\,\%pt) for the smallest train set (\textbf{\#train=50}).
For larger train sets, the gap is 14-15\,\%pt.
Using Bayesian analysis (see equation~\ref{e:bayesian_success_rates}), which effectively incorporates both the experimental success rate and the sample size into one figure of merit, we can make the following statement. 
According to our data, it is at least 84.7\,\% certain that an instrumented policy outperforms a vision-only policy. 
However, an instrumented policy cannot easily be deployed in the field: we do not envision every clothes hanger in the bedroom closet having embedded sensors.
In this work, we used the instrumented policy $\Pi^{180}_{\textrm{instr}}$ to gather additional data, part of which was specifically targeted to failures of the vision-only policy $\Pi^{180}_{\textrm{vis}}$.
A new vision-only policy $\Pi^{237}_{\textrm{vis+}}$ trained on this enhanced data set showed a performance similar to $\Pi^{180}_{\textrm{instr}}$---it makes sense that the student does not outperform the teacher---and outperformed $\Pi^{180}_{\textrm{vis}}$ by 12\,\%pt.
In the Bayesian interpretation (equation~\ref{e:bayesian_success_rates}), we are 85.3\,\% certain that $\Pi^{237}_{\textrm{vis+}}$ outperforms $\Pi^{180}_{\textrm{vis}}$.

Apart from success rates, we also note substantial differences in the policy failure modes and behaviours.
For $\Pi^{50}_{\textrm{vis}}$, 5 out of 18 failures occurred when the left gripper dropped the T\nobreakdash-shirt at a point where the hanger insertion was not completed (\textbf{Drop\,[1st insertion]} + \textbf{Drop\,[2nd insertion]}).
For $\Pi^{100}_{\textrm{vis}}$ 7 out of 16 failures are \textbf{Drop\,[1st insertion]}: the policy seems to execute robot movements with little regard for the state of the T\nobreakdash-shirt (see Fig.~\ref{fig:not_aware_n100}). 
This indicates little task awareness of the vision-only policy at small training set sizes.
$\Pi^{50}_{\textrm{instr}}$ had three \textbf{Drop\,[2nd insertion]} failures out of 13, while no more \textbf{Drop} failures occurred for $\Pi^{100}_{\textrm{instr}}$.
This tendency of getting stuck rather than failing catastrophically opens the door for more effective human takeovers during interactive IL like~\cite{ingelhag2025realtimeoperatortakeovervisuomotor, kelly2019hgdaggerinteractiveimitationlearning, dai2024racerrichlanguageguidedfailure}.

Importantly, the \textbf{Drop} failures of $\Pi^{50}_{\textrm{instr}}$ were caused by the outer sensor of the second hanger leg (*1 in Fig.~\ref{fig:clothes_hanger_annotated}) being covered by a folded sleeve of the T\nobreakdash-shirt (see Fig.~\ref{fig:2nd_insert_drop}).
This shows how strongly the $\Pi_{\textrm{instr}}$ policies rely on the instrumentation data.
Additional evidence is shown in Fig.~\ref{fig:instrumentation_cue}: $\Pi^{50}_{\textrm{instr}}$ misses the first insertion and realises it is not ready to lift.
When we manually cover the sensor that should be in the T\nobreakdash-shirt at this point, the visual cues are ignored and the policy proceeds to lift.
This is an important and non-trivial result underpinning the potential of instrumentation: a black-box IL policy can recognise the importance of instrumentation signals without explicit guidance.

These results favour instrumentation for IL, but we must keep in mind the question of whether the effort spent implementing instrumentation had not better been redirected to collecting additional teleop demonstrations~\cite{mirchandani2024thinkscaleautonomousrobot}.
In this case study, collecting an additional 57 teleop demonstrations would take less than 1.5\,h, which is less than the development time of the instrumentation system.
However, while the integration of infrared range sensors is specifically designed for the task of clothes hanger insertion, the modular readout PCB has already proven to be applicable to multiple tasks~\cite{smarttextile, ursquee}.
Hence, expanding this open-source platform of instrumentation hardware can accelerate development of instrumented objects and enable instrumented learning.

Even still, instrumentation requires a certain data scale to be beneficial.
Building intuition towards where this turning point lies is a key point of interest for future research.
However, before we scale up robot learning experiments to larger datasets, it pays to investigate alternative methods of leveraging instrumentation data during training, while obtaining a policy that can be deployed without instrumentation input.
Several methods are proposed in section~\ref{s:conclusion}.

Furthermore, we believe that the performance gap between instrumented and vision-only policies may increase with less constraints on the task, i.e. more types of clothes hanger, more T\nobreakdash-shirts, more variation in the position of the T\nobreakdash-shirts.
These variations highly impact the vision modality, while the instrumentation modality is relatively unbothered: even with varying IR absorption between different coloured T\nobreakdash-shirts, the difference in sensor readings between covered and uncovered states is tremendous.

\section{Conclusion \& Future Work} \label{s:conclusion}
This work presents foundational results uncovering the potential of instrumentation for IL through a case study focusing on clothes hanger insertion. 
We developed an instrumented clothes hanger by integrating IR reflective range sensors, which provide a strong signal when the clothes hanger is covered compared to when it is not.
This way, the instrumentation provides a clear indication of task progression, which cannot be directly inferred from other modalities such as vision or, potentially, tactile sensing.
A dataset of 180 teleop demonstrations of the hanger insertion task was collected.
Three different subsets of this dataset were defined, and for each subset two Diffusion Policy models were trained: one including instrumentation data in the state vector, a second excluding instrumentation data.
First and foremost, we found that instrumented policies recognise the importance of instrumentation signals without explicit guidance, even prioritising them above visual cues.
Second, instrumented policies outperform their vision-only counterparts by 14-25\,\%pt.
Lastly, we showed that adding rollouts from an \enquote{expert} instrumented policy to the teleop dataset and training a \enquote{student} vision-only policy on this enhanced dataset increases the performance of the vision-only student to the level of the instrumented expert.

A critical note is that in this case study the performance gap between instrumented and vision-only policies is not large enough to warrant the effort spent on implementing instrumentation.
However, the readout PCB used in this work has already proven valuable for multiple tasks, so further development of this open-source instrumentation platform can enable efficient instrumented learning.
Furthermore, we expect the performance gap to increase when allowing for more variation in the task.
Finally, using an instrumented policy to obtain expert rollouts is just one possible way to leverage instrumentation to achieve a more performant vision-only policy.
We outline several approaches to be developed in future work:
\begin{itemize}
    \item \textbf{Soft sensor:} An additional, specialised model is trained to predict the sensor values from the camera images. At inference time, the predictions of this \enquote{soft sensor}~\cite{softsensor2023} are used for the instrumentation inputs of the action policy. This approach effectively decomposes the task in (1) explicit state estimation and (2) action selection given a certain state.
    \item \textbf{Pretraining:} First train a ResNet18 model to predict the sensor values given the camera images, and then use it as the vision backbone for the Diffusion Policy network. By pretraining on the sensor data, we aim to capture essential information about the goal and task progression in the backbone network, which provides the action policy with a head-start for training. Previous work has shown that pretraining a visuo-tactile encoder and using it as the vision backbone for a vision-only IL policy can drastically increase performance~\cite{vital2024}.
    \item \textbf{Additional training loss:} We train the network as a whole to predict both robot actions and the sensor data, either by using a concatenation of the sensor values and teleop actions as training labels, or by adding a second prediction head and combining its loss with that of the robot action prediction head. In essence, this forces the model to explicitly predict whether or not the clothes hanger is covered, which is a crucial indicator of task progression.
\end{itemize}
Once we have a broad understanding of how instrumentation data can be leveraged to speed up robot learning, we can conduct experiments at larger data scales and discover when instrumented learning can reduce overall human effort in obtaining functional robot manipulation policies.

\section{Acknowledgment}

GPT-5 was used to write the first draft of the abstract. GPT-5 and Writefull
were used sparingly throughout the manuscript to improve specific sentences and wordings.

\balance
\bibliographystyle{IEEEtran}
\bibliography{references.bib}

\end{document}